\pdfoutput=1 
\documentclass[10pt,twocolumn,letterpaper]{article}
\usepackage[final]{cvpr}
\usepackage{times}
\usepackage{epsfig}
\usepackage{graphicx}
\usepackage{amsmath}
\usepackage{amssymb}
\usepackage{booktabs}

\usepackage[pagebackref,breaklinks,colorlinks]{hyperref}

\usepackage[capitalize]{cleveref}
\crefname{section}{Sec.}{Secs.}
\Crefname{section}{Section}{Sections}
\Crefname{table}{Table}{Tables}
\crefname{table}{Tab.}{Tabs.}

\usepackage[utf8]{inputenc}
\usepackage{xcolor}
\usepackage{endnotes,color,url,paralist, multirow,float}
\usepackage{verbatim}
\usepackage{epstopdf}
\usepackage{comment}
\usepackage{bm}
\usepackage[]{algorithm2e}
\usepackage{wrapfig}
\usepackage{afterpage}
\usepackage{makecell}
\usepackage{balance}
\usepackage{ctable}
\usepackage{tabularx}
\DeclareMathOperator*{\argmin}{argmin}
\usepackage{mathtools, nccmath}
\usepackage{gensymb}
\usepackage{url}
\usepackage{multirow}
\usepackage{xpatch}
\usepackage{soul}
\usepackage{enumitem}

\def\cvprPaperID{10467} 
\def\confName{CVPR}
\def\confYear{2023}

\begin{document}
\title{ViFi-Loc: Multi-modal Pedestrian Localization using GAN with Camera-Phone Correspondences}
\author{Hansi Liu\\
Rutgers University\\
{\tt\small hansiiii@winlab.rutgers.edu}
\and
Kristin Dana\\
Rutgers University\\
{\tt\small kristin.dana@rutgers.edu}
\and
Marco Gruteser\\
Rutgers University\\
{\tt\small gruteser@winlab.rutgers.edu}
\and
Hongsheng Lu\\
Toyota Motor North America\\
{\tt\small hongsheng.lu@toyota.com}
}

\maketitle

\begin{abstract}
    In Smart City and Vehicle-to-Everything (V2X) systems, acquiring pedestrians' accurate locations is crucial to traffic safety. Current systems adopt cameras and wireless sensors to detect and estimate people's locations via sensor fusion.
    Standard fusion algorithms, however, 
    become inapplicable when multi-modal data is not associated. For example, pedestrians are out of the camera field of view, or data from camera modality is missing.
    To address this challenge and produce more accurate location estimations for pedestrians, 
    we propose a Generative Adversarial Network (GAN) architecture. During training, it learns the underlying linkage between pedestrians' camera-phone data correspondences.
    During inference, it generates refined position estimations based only on pedestrians' phone data that consists of GPS, IMU and FTM.
    Results show that our GAN produces 3D coordinates at 1 to 2 meter localization error across 5 different outdoor scenes.
    We further show that the proposed model supports self-learning. The generated coordinates can be associated with pedestrian's bounding box coordinates to obtain additional camera-phone data correspondences. This allows automatic data collection during inference.
    After fine-tuning on the expanded dataset, localization accuracy is improved by up to 26\%.
\end{abstract}

\section{Introduction}
\begin{figure}[]
    \centering
    \includegraphics[width=0.8\columnwidth]{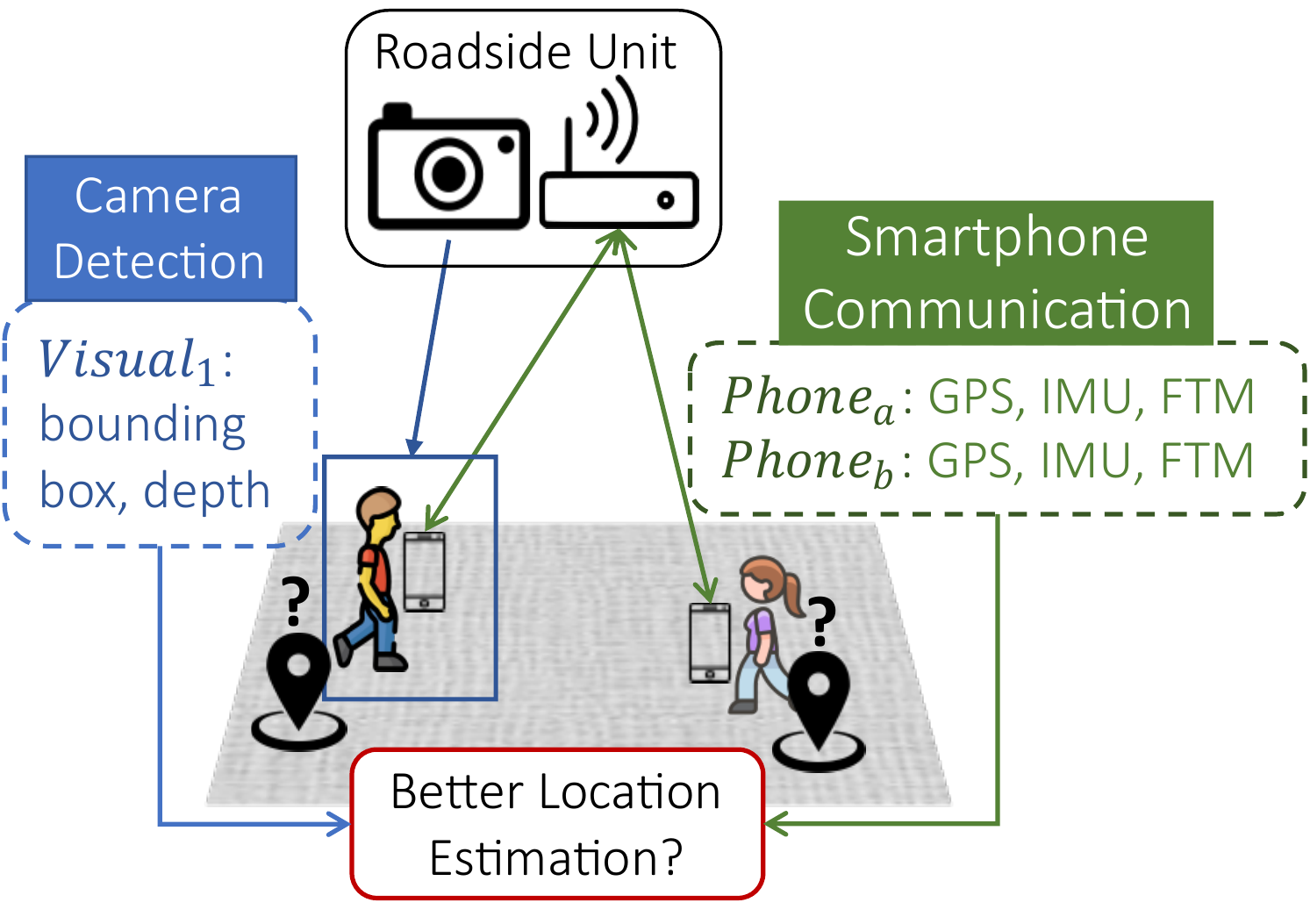}
    
    \caption{Motivation. The roadside unit collects multi-modal data by ``seeing'' a group of pedestrians via camera and ``hearing'' a group of phone devices via wireless communication. Can we provide pedestrians with accurate location estimations leveraging the multi-modal data that is not necessarily associated?}
    \label{fig:motivation}
    \vspace{-6mm}
\end{figure}

To enhance traffic and pedestrian safety, Vehicle-to-Everything (V2X) systems leverage vision and wireless sensing to estimate and share traffic participants' locations.
However, the estimated pedestrian locations may not be accurate because each sensing modality has its own limitations. Camera RGBD sensing, while providing accurate depth information of the detected persons, is limited to non-line-of-sight (NLOS) scenarios and suffers from drastic illumination changes. Wireless sensing such as Fine-time-measurement (FTM)~\cite{80211standard} is more robust to NLOS conditions, but its ranging performance degraded by multi-path and shadowing in complex environment. 

It is desirable to combine both modalities to achieve better localization for pedestrians. The state-of-the-art multi-modal sensor fusion algorithms usually provide a state estimation which is more accurate than the measurement from each single modality. For example, fusing camera depth measurement with FTM allows ranging measurement to be more accurate; Fusing GPS with IMU allows fine-grained localization that is robust to sensor noise and drifting.

These standard data fusion approaches, however, are applicable only under the condition that measurements from both modalities are available and data association is known. If a subject's multi-modal data is not correctly associated, then the fused measurement would be meaningless. Moreover, measurements of a pedestrian might not always include both modalities. For a situation depicted in Figure~\ref{fig:motivation}, when a person is out of the camera view or the detection algorithm fails to detect her and only phone data is available, it would be infeasible to leverage data from camera modality to improve the performance of wireless ranging and localization. This challenge motivates us to come up with a solution that accurately estimates a person's location using information from both camera and phone modalities and not depending on pre-computed data association. 

Considering a pedestrian's camera data contains bonding box coordinates and phone data includes GPS, IMU measurements and FTM, we can view the localization task as \textit{refining} or \textit{correcting} a person's raw GPS data using his camera RGBD data, IMU measurements and FTM.
An important intuition lies in the fact that GPS localization error, affected by satellite constellation, tends to be correlated for pedestrians in the vicinity within a period of time ~\cite{rostami2019light}. Thus, if we can learn a cross-modal mapping between a group of pedestrians' existing camera-phone data correspondences, we will be able to use the same mapping to localize other pedestrians in the same area, even if they are out of the camera field of view, by translating their phone data into local camera spatial coordinates.
We propose a cross-modal Generative Adversarial Network (GAN) architecture that learns the linkage between the a person's camera modal measurement and phone modal measurement. 
During training, measurements of a pedestrian's multi-modal data within a time window will be fed into the network. Measurements from phone modality includes FTM, IMU data and smartphone GPS readings; Measurements from camera modality include bounding box centroids and depth measurement. Although phone measurement and camera measurement are not directly comparable, they both describe and encode the same pedestrian's walking trajectory and gait information. To reflect this linkage, the network extracts representative features from input measurements and enforce the extracted features to be close to each other in the hyper-space. Then a decoder is applied to the feature vector from phone modality to generate a location estimation with respect to the camera coordinate frame. The generated coordinate will be examined by a discriminator to ensure that the produced coordinate is within the distribution of true locations.
During inference, the proposed network is capable of generating a person's estimated coordinates with respect to the local camera coordinate frame based on the person's phone measurements. 
Our results show that the proposed GAN architecture generates spatial coordinates at an average localization error of 1.5 m, which significantly outperforms the baseline approach, which fuses smartphone's original GPS measurement with FTM and has an average localization error of 5.3 m.

Training the proposed network requires multi-modal data correspondences from one or more experiment participants. Collecting and labeling multi-modal data from multiple pedestrians in larger scale is time and resource consuming.
Here we further propose a semi-supervised approach that leverages the network output to automatically produce more multi-modal data correspondences to facilitate larger scale training. 
Upon obtaining the estimated locations from pedestrian's phone measurement, we associate the GAN produced coordinates with the existing bounding box centroids 3D coordinates. Since each GAN produced location corresponds to a pedestrian's phone measurement, associating the GAN produced location with camera modality measurement is essentially acquiring additional camera-phone data correspondences. 
The semi-supervised approach allows us to easily obtain large scale reliable training data without dedicated data collection and manual labeling.

To the best of our knowledge, we are the first to apply GAN to generate camera modality location measurements from phone measurements. Unlike existing works that use GAN to generate synthesized images~\cite{bao2017cvae, lu2018image, han2018gan}, texts~\cite{gorti2018text} or WiFi signals~\cite{shi2020generative, shi2019generative}, we are focusing on generating 3D locations from wireless ranging measurements, IMU measurements and camera bounding box coordinates and depth. 


\section{Background and related work}
\textbf{Vison-based localization} \quad There is a lot of related work on pedestrian localization. These work can be categorized based on sensors' types and modalities. In vision domain, localization can be achieved by camera or other optical sensors such as lidar. Using RGBD or 3D pointcloud information and state-of-the-art human detectors, a person's spatial location can be estimated. Off-the-shelf RGBD cameras with person detection and localization functionality such as ZED and RealSense offers with a depth accuracy of 1\% of the distance in the near range to 9\% in the far range within 20 meters \cite{ZED_accuracy, Realsense_accuracy}.

\textbf{GPS} \quad In smartphone wireless sensors domain, one of the typical positioning techniques is GPS. Different GPS solutions, with different chip set configurations and services, provide a varying range of localization granularity from meter level to sub-centimeter level. GPS-enabled smartphones are typically accurate within a 5 meter radius under open sky \cite{GPS_gov}. However, their performances are usually affected and degraded by factors including satellite constellation, poor weather condition, environment variation and multipath due to tall building, bridges and trees. Pocket-size GPS receivers with moderate prices offer positioning accuracy around tens of meters~\cite{misra1999gps}. More specifically, the study in \cite{zhang2018quality} suggests that  the observation quality of android smartphone GNSS observations is difficult to achieve meter-level accuracy if using only pseudo-range observations. While survey grade GPS equipment achieves sub-centimeter accuracy, it usually requires specialized equipment and expensive configurations with extra subscription services such as Real Time Kinematics (RTK) and Real Time Differential~\cite{gao2021raw}. 

\textbf{WiFi localization} \quad Another major category of studies on localization focuses on WiFi signals. Received Signal Strength Indicator (RSSI) can be used in fingerprinting ~\cite{cheng2005accuracy} and trilateration~\cite{ismail2019rssi}. WOLoc~\cite{wang2017woloc} achieves meter-level accuracy in outdoor environment through semi-supervised manifold learning with multiple access point's RSSI.
More recently, WiFi Fine Time Measurements (FTM)~\cite{80211standard} has been extensively explored in localization tasks .
\cite{ibrahim2018verification} confirms that the FTM protocol can achieve meter-level accuracy in open space environments although degrades in high multipath environments.

\textbf{Inertial aided localization} \quad
Inertial Measurement Units (IMU) are often adopted as auxiliary sensors due to their easy accessibly and cheap prices. IMU can provide kinemetic information at higher sample rate than GPS and WiFi message exchange rate. Localization based on IMU dead-reckoning alone, however, suffers from cumulative error in long term. Standard approaches to reduce the cumulative error include filtering techniques that incorporates IMU with GPS and WiFi measurements. For example, ~\cite{cheng2014seamless} fuses WiFi RSSI fingerprinting, GPS and IMU using an Extended Kalman Filter.
Wi-Go~\cite{ibrahim2020wi} achieves outdoor vehicular localization error of 1.3 m median and 2.9 m 90-percentile. It simultaneously tracks vehicles and locates WiFi access points by fusing WiFi FTM, GPS, and vehicle odometry information using a particle filter. 

\textbf{Multi-modal sensor fusion} \quad Vision-based localization and wireless-based localization have complementary characteristics. Camera sensing provides more accurate spatial information in the near field through RGBD sensing, but they suffer from occlusion, appearance and illumination variation; wireless sensing, on the other hand, can work in non-line-of-sight and poor illumination conditions. But its ranging performance can be degraded by complex environments with multi-path and shadow fading. Combining vision and wireless sensing in a localization system has gain more attention recently as it combines both modalities' advantages. Related work in the cross field includes  Simultaneous Localization and Mapping (SLAM), where a mobile agent relies on vision and wireless data to locate itself while creating a representation of the surrounding. SLAM can be achieved using vision only~\cite{mur2015orb}, vision+IMU~\cite{campos2021orb, qin2018vins}, WiFi+IMU~\cite{ibrahim2020wi, arun2022p2slam}, etc.. Compared with traditional Filtering approaches such as EKF and particle filters, Bundle Adjustment, pose graph or factor graph optimization~\cite{loeliger2004introduction, dellaert2012factor} provides better performance in larger scale. 

The above-mentioned sensor fusion approaches has a limitation that vision and wireless data need to be available at the same time during inference. In our task, a person's camera data could be unavailable due to limited field of view. As a result, typical Kalman filter or SLAM approaches become inapplicable. A novel approach is needed to fully exploit pedestrians' camera data and phone data.


\section{Multi-modal location estimation}
\begin{figure}[]
    \centering
    \includegraphics[width=0.4\textwidth]{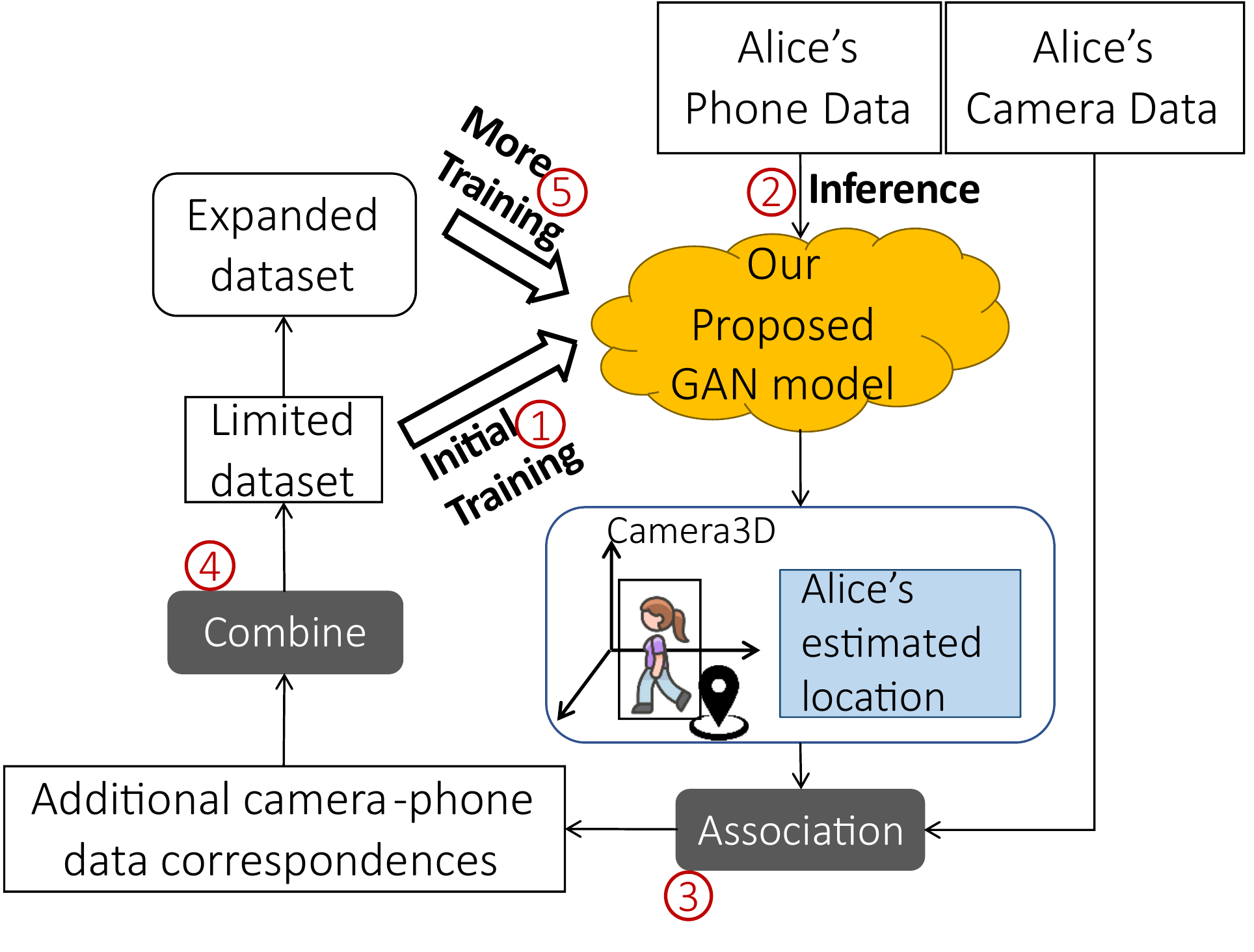}
    \caption{Method overview. A GAN model is trained on initial limited dataset that contains pedestrians' camera-phone data correspondences. During inference, the model generates a location estimation for the user based only on her phone data. The generated coordinates can be use to associate with camera bounding box coordinates to produce additional camera-phone data correspondences, which allows the dataset to expand automatically. More training and fine-tuning on the expanded dataset improve the model's localization accuracy.}
    \label{fig:overview}
    \vspace{-4mm}
\end{figure}

Figure~\ref{fig:overview} presents an overview of our methodology. 
The model is first trained with a manually labeled dataset that consists of camera-phone data correspondences of multiple pedestrians. During inference, the network produces location estimations based only on pedestrians' phone data. The produced coordinates are then associated with bounding box coordinates from the camera modality. Then the associated data correspondences are combined with the original dataset to form a larger scale training set that can be used to further train the network. This feedback loop enables our network to achieve self-learning --- using the network's output to produce more training data during inference.

Next, we introduce our GAN architecture, how it is trained and how the generated locations is associated with camera observations to obtain more training data.
\subsection{GAN Architecture}
\begin{figure*}[!h]
    \centering
    \includegraphics[width=0.90\textwidth]{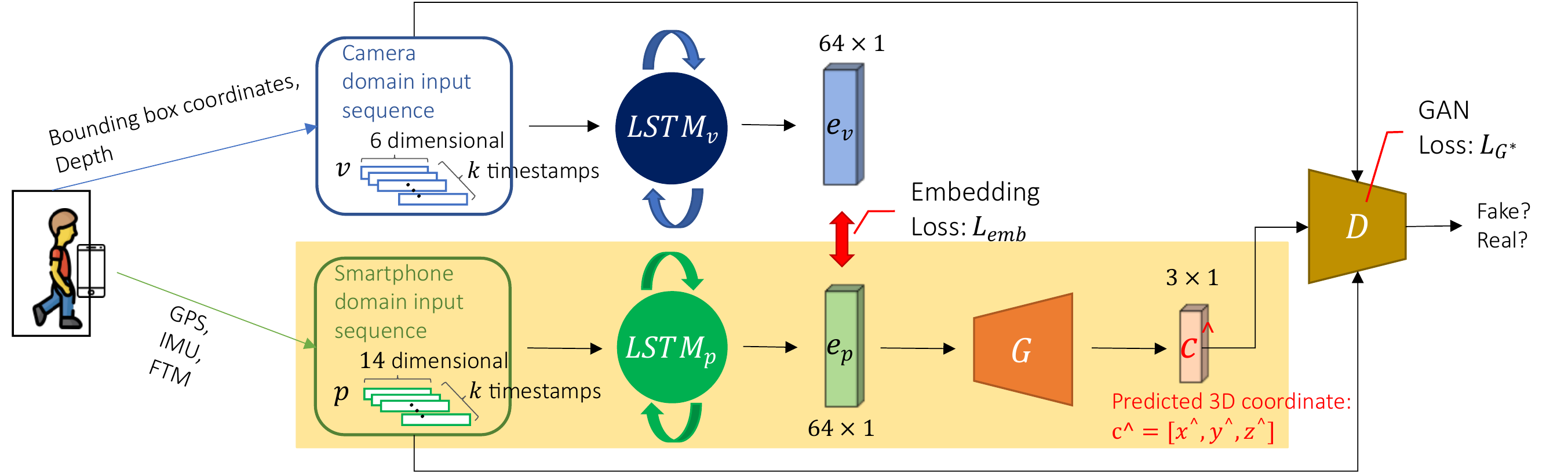}
    \caption{Proposed GAN architecture. The input includes pedestrians' camera-phone data correspondences. Camera domain input consists of the person's bounding box centroid and depth information. Phone domain input contain FTM range, standard deviation, 9-axis IMU and GPS coordinates w.r.t. local camera 3D coordinate frame. The input goes through two independent bi-directional LSTM units. The output embeddings $\bm{e}_v$ and $\bm{e}_p$ encode spatial and temporal cues of the person's camera data and phone data. They are constrained by the embedding loss becuase they come from the same pedestrian. A Generator $\mathcal{G}$ renders the phone modality embedding $\bm{e}_p$ into a coordinate $\bm{c}^\wedge$. A discriminator $\mathcal{D}$ is used to examine whether $\bm{c}^\wedge$ is genuine or fake. The detailed configurations of $\mathcal{G}$ and $\mathcal{D}$ are listed in Table~\ref{table:network_G_config} and~\ref{table:network_D_config}. Training the network uses data from both modalities. Inference only requires phone data as input, as shown in yellow shaded region.}
    \label{fig:GAN_arch}
    \vspace{-2mm}
\end{figure*}

Figure \ref{fig:GAN_arch} shows the proposed GAN architecture. The network takes as input a pedestrian's sequential multi-modal data within a time window $k$. For every timestamp $t$, the vision data $\bm{v}_t$ takes the form
\begin{equation}
\small
    \bm{v}_t = 
    \begin{bmatrix}
         d, & x, & y, & X, & Y, & Z
    \end{bmatrix}
    \in \mathbb{R}^6,
\end{equation}
where $d$ is the depth value of the pedestrian's bounding box centroid; [$x$, $y$] is pixel coordinate of the bounding box centroid; [$X$, $Y$, $Z$] is the bounding box centroid's 3D coordinate with respect to the camera coordinate frame.

For wireless data the input at timestamp $t$ takes the form 
\begin{equation}
\small
    \bm{p}_t = 
    \begin{bmatrix}
         r_{\text{ftm}}, & \textit{std}_{\text{ftm}}, & \textit{Acc}, & \textit{Gyr}, & \textit{Mag}, & \textit{GPS} 
    \end{bmatrix}
    \in \mathbb{R}^{14}.
\end{equation}
It contains FTM range $r_{\text{ftm}}$, FTM standard deviation $\textit{std}_{\text{ftm}}$, 9-axis IMU data (accelerometer $[x_{\text{acc}}, y_{\text{acc}}, z_{\text{acc}}]$, gyroscope $[x_{\text{gyr}}, y_{\text{gyr}}, z_{\text{gyr}}]$, and magnetometer $[x_{\text{mag}}, y_{\text{mag}}, z_{\text{mag}}]$) as well as GPS coordinates $[X_{\text{gps}}, Y_{\text{gps}}, Z_{\text{gps}}]$ with respect to the local camera coordinate frame. 

The synchronized vision and wireless sequential data are rendered by two independent bi-directional LSTM modules into feature embeddings $\bm{e}_v$ and $\bm{e}_p$, which contains spatial and temporal cues of the person's camera modality input and phone modality input.
We adopt LSTM units as feature extractors for the multi-modal input considering they offer significant advantages over other vanilla multi-layer network architectures when extracting features from
sequential or time-series data.
Because these feature vectors represents the same pedestrian, we use the Embedding Loss to force them to be close to each other in the high-dimensional space. 
The Embedding Loss takes the form $ L_{\textit{emb}} = \| \bm{e}_v - \bm{e}_p \|_2$.
It ensures the linkage between vision modality and wireless modality.

Next, the wireless modality feature vector $\mathbf{e}_p$ goes through a generator $\mathcal{G}$ that consists of a series of fully connected layers, batchnorm layers and dropout units. The detailed architecture is listed in Table~\ref{table:network_G_config}. The generator renders the feature vector $\mathbf{e}_p$ into a coordinate $\bm{c}^\wedge$ in the camera's 3D local coordinate frame.
The generated coordinate along with the network input are then examined by a discriminator $\mathcal{D}$ whose detailed architecture is listed in Table~\ref{table:network_D_config}.
$\mathcal{G}$'s purpose is to generate valid coordinates that is within the distribution of true location;
$\mathcal{D}$'s purpose is to stringently discriminate or examine if a generated coordinate is good enough, i.e., within the distribution of the pedestrian's true coordinates.
The output of the discriminator $\mathcal{D}$ is 0 or 1 indicating whether the examined coordinate is unrealistic (fake) or realistic (true).
We use the Generative loss to train the generator and discriminator. 
The Generative Loss $L_{G^*}$ takes the form
\begin{equation}
\small
        L_{G^*} = \min_{\mathcal{G}}\max_{\mathcal{D}} L_{\textit{LSGAN}}(\mathcal{G}, \mathcal{D}) +  g(\bm{c}_{\text{gnd}}, \bm{c}^\wedge),
\end{equation}
where $L_{\textit{LSGAN}}(.)$ is the standard Least Squares GAN Loss~\cite{mao2017least}
\begin{equation}
\begin{aligned}
\small
    L_{\textit{LSGAN}}(\mathcal{G}, \mathcal{D}) = \quad & \mathbb{E}[(\mathcal{D}(\bm{p}, \bm{v}, \bm{c}_{\text{gnd}})-1)^2] \\
    & + \mathbb{E}[\mathcal{D}(\bm{p}, \bm{v}, \mathcal{G}(\bm{e}_p))^2];
\end{aligned}
\end{equation}
$g(.)$ is the regularization term
\begin{equation}
    g(\bm{c}_{\text{gnd}}, \bm{c}^\wedge) = |\bm{c}_{\text{gnd}} - \bm{c}^\wedge| + \| \bm{c}_{\text{gnd}} - \bm{c}^\wedge \|_2,
\end{equation}
It penalizes the reconstruction loss of the predicted coordinate $\bm{c}^\wedge$ and the ground-truth coordinate $\bm{c}_{\text{gnd}}$.
The total loss is the sum of the Embedding Loss and the Generative Loss
\begin{equation}
    L =  L_{\textit{emb}} +  L_{G^*}.
\end{equation}



During training, both $\mathcal{G}$ and $\mathcal{D}$ will improve as they combat with each other. $\mathcal{G}$ will be better at predicting valid coordinates, and $\mathcal{D}$ will be better at determining fake generated coordinates. 
Eventually, an equilibrium is achieved during training, and the generated coordinates will be used as location estimations. 

We implement the network architecture using PyTorch \cite{NEURIPS2019_9015} -- the detailed configurations of the network layers are listed in Table~\ref{table:network_G_config} and \ref{table:network_D_config}. We train the network with an NVIDIA 1080-Ti GPU for 200 epochs with batch size of 32, learning rate 0.001 (0.0001 after 100 epochs).

\begin{table}[]
\centering
\footnotesize
    \begin{tabular}{|c||c|c|}
    \hline
    Input & $\bm{v} \in \mathbb{R}^{6 \times 10}$ & $\bm{p} \in \mathbb{R}^{ 14 \times 10}$ \\ \hline
    \begin{tabular}[c]{@{}c@{}}Feature \\ extractor\end{tabular} & \begin{tabular}[c]{@{}c@{}}$\text{LSTM}_v$ (6, 64)\end{tabular} & \begin{tabular}[c]{@{}c@{}}$\text{LSTM}_p$ (14, 64)\end{tabular} \\ \cline{1-3}
    
    \begin{tabular}[c]{@{}c@{}}Extracted \\ feature\end{tabular} &
    \begin{tabular}[c]{@{}c@{}}$e_v \in \mathbb{R}^{64 \times 1}$\end{tabular} & \begin{tabular}[c]{@{}c@{}}$e_p \in \mathbb{R}^{64 \times 1}$\end{tabular} \\ \cline{1-3}
    \multicolumn{2}{|c||}{\multirow{8}{*}{\begin{tabular}[c]{@{}c@{}}Layers \\ of $\mathcal{G}$\end{tabular}}} 
    & \multicolumn{1}{c|}{\begin{tabular}[c]{@{}c@{}}FC1 (64, 64),  BatchNorm1D \\ Leaky-ReLU, Dropout\end{tabular}} \\ \cline{3-3} 
    \multicolumn{2}{|c||}{} &\multicolumn{1}{c|}{\begin{tabular}[c]{@{}c@{}}FC2 (64, 64),  BatchNorm1D \\ Leaky-ReLU, Dropout\end{tabular}} \\ \cline{3-3} 
    \multicolumn{2}{|c||}{} &\multicolumn{1}{c|}{\begin{tabular}[c]{@{}c@{}}FC3 (64, 64),  BatchNorm1D \\ Leaky-ReLU, Dropout\end{tabular}} \\ \cline{3-3} 
    \multicolumn{2}{|c||}{} &\multicolumn{1}{c|}{\begin{tabular}[c]{@{}c@{}}FC4 (64, 32), BatchNorm1D \\ Leaky-ReLU\end{tabular}} \\ \cline{3-3} 
    \multicolumn{2}{|c||}{} &\multicolumn{1}{c|}{\begin{tabular}[c]{@{}c@{}}FC5 (32, 3)\end{tabular}} \\ \hline
    \multicolumn{2}{|c||}{Output} & \multicolumn{1}{c|}{$\bm{c}^{\wedge} \in \mathbb{R}^{3 \times 1}$} \\ \hline
    \end{tabular}
    \caption{Detailed configuration of $\mathcal{G}$.}
  \label{table:network_G_config}
\end{table}

\begin{table}[]
\centering
\footnotesize
    \begin{tabular}{|c||c|c|c|}
    \hline
    Input & $\bm{v} \in \mathbb{R}^{6 \times 10}$ & $\bm{p} \in \mathbb{R}^{ 14 \times 10}$ & \multirow{1}{*}{\begin{tabular}[c]{@{}c@{}}$\bm{c}^{\wedge} \in \mathbb{R}^{3 \times 1}$\end{tabular}}\\ \cline{1-4}
    \multirow{6}{*}{\begin{tabular}[c]{@{}c@{}}Layers \\ of $\mathcal{D}$\end{tabular}}  & \begin{tabular}[c]{@{}c@{}}$\text{LSTM}_v$ (6, 8)\end{tabular} & \begin{tabular}[c]{@{}c@{}}$\text{LSTM}_p$ (14, 8)\end{tabular} & \\ \cline{2-4}
    & \multicolumn{3}{c|}{\begin{tabular}[c]{@{}c@{}c@{}}FC1 (19, 8),  BatchNorm1D \\ Leaky-ReLU\end{tabular}}  \\ \cline{2-4}
     & \multicolumn{3}{c|}{\begin{tabular}[c]{@{}c@{}c@{}}FC2 (8, 4),  BatchNorm1D \\ Leaky-ReLU\end{tabular}} \\ \cline{2-4} 
     & \multicolumn{3}{c|}{\begin{tabular}[c]{@{}c@{}c@{}}FC3 (4, 1)\end{tabular}} \\ \cline{1-4} 
    Output & \multicolumn{3}{c|}{$d \in \mathbb{R}$} \\ \hline
    \end{tabular}
    \caption{Detailed configuration of $\mathcal{D}$. }
    \vspace{-6mm}
  \label{table:network_D_config}
\end{table}

\subsection{Self-learning with association}

From a practical perspective, to train a network that can work at large scale and various scenarios, the more training samples there is, the better. 
Training the proposed network architecture requires a large amount of labeled vision-phone data correspondences. Obtaining sufficient data correspondences requires collecting multi-modal data from multiple pedestrians in various outdoor scenarios. Besides, a lot of extra effort is needed to determine and label the vision-phone correspondences from the collected multi-modal data so that they can be used in the training process. 
Although there exists available labeled muti-modal dataset that are of reasonable scale for us to initially train the proposed network, it would be challenging to improve the network's performance by training on larger scale dataset.


To address this challenge, we propose a self-training mechanism for our network to acquire more vision-phone data correspondences during inference, we associate pedestrians' camera domain coordinates with our GAN output coordinates. Since the input of the GAN during inference is pedestrians' phone data, by solving the association problem we are essentially finding the correct vision-phone correspondences in the test data. 
The network is first trained with a limited portion of the labeled data correspondences. During the test phase, suppose at a timestamp in the test data there are $M$ camera detected bounding boxes and $N$ available phone data sequences. Using RGBD information we can obtain $M$ camera 3D coordinates $\{p^{\text{camera}}\}$; using our GAN to perform inference on these phone data sequences we have $N$ generated coordinates $\{p^{\text{phone}}\}$ that are with respect to the camera local 3D coordinate frame. 
Because the GAN is trained to produce realistic coordinates that are close to the pedestrian's true camera coordinates, for a true camera-phone data correspondence, the distance between its camera coordinates and its GAN generated coordinates should be smaller than that of non-correspondences. 

Using this huristic, we choose the camera observation whose bounding box coordinate has the smallest Euclidean Distance to the GAN produced coordinate as the associated identity from camera modality for every identity in the phone modality:
\begin{align}
\small
    \text{AssociatedID}_i = \argmin_{j \in [1, M]} \|p^{\text{phone}}_i - p^{\text{camera}}_j\|_2.
\end{align}
In this way, we obtain good quality camera-phone correspondences as additional training samples without dedicated effort of data collection or manual labeling. Then we combine the associated data with the initial labeled data to form a larger dataset that can be used to retrain or fine tune the network. The feedback loop in Figure~\ref{fig:overview} indicates that the pipeline of train-association-retrain can be executed in multiple iterations, with each iteration the association can bring in more new data correspondences. This allows the network to evolve on its own after initially trained with a limited amount of labeled data correspondences.

\section{Evaluation}
\textbf{Dataset}\quad
We adopt the multimodal dataset in Vi-Fi~\cite{liuvi} to train the network. The dataset contains pedestrian's camera data and their smartphone's wireless measurement. Camera data includes bounding box centroid and depth measurements; wireless measurements contains smartphone GPS readings, FTM ranging and IMU measurements. 
The setup of dataset collection contains a roadside unit (RSU) that consists of an RGBD camera and WiFi access point that are placed together.
A mounted Stereolabs ZED2~\cite{ZED} (RGB-D) camera is set at the height of 2.4--2.8 meters with a proper field of view to record video at 3fps, which collects depth information from 0.2m to 20m away from the camera. The smartphones are set to exchange FTM messages at 3 Hz frequency with a Google Nest WiFi Access Point anchored besides the camera. Each smart phone also logs its IMU sensor data at 50 Hz and GPS readings at 1 Hz (in Dataset B only).  The smartphones and camera are connected to the Internet to achieve coarse synchronization using network time synchronization. 

The dataset contains in total 79 3-minute video sequences across 5 outdoor scenarios. We randomly choose 1 sequence from each scenario and use the vision-phone data correspondences from multiple pedestrians to form the test set. We use the the data from the rest 74 sequences to form the training set. Each data sample contains a pedestrian's multi-modal data within a 3-second time window. The total number of training sample is 110141 and the total number of testing sample is 6951.


\textbf{Baseline} \quad 
Since the problem of cross-modal coordinate generation has not been specifically addressed in the literature, there are no off-the-shelf model architectures to compare against. Common solutions to localization adopt filter-based approaches to fuse measurements from multiple sources to obtain better estimations. In our context, however, the multi-modal data from pedestrians are not associated. When the camera data is not available, we can only rely on  pedestrians' phone data to estimate their locations. Therefore, we choose phone GPS readings and a particle filter that fuses GPS with FTM as our baselines. The phone GPS data is obtained by an Android API function that returns the standard GNSS observations~\cite{LocationManager}. The particle filter approach corrects each GPS measurement with the RSU's FTM ranging information.

\subsection{Localization accuracy}
The coordinates from the camera modality input are with respect to the local camera 3D coordinate frame; the coordinates from phone modality input are with respect to the world's GPS system. To train our network and evaluate the accuracy of the predicted coordinates, the coordinates from multi-modal input need to be in the same reference system. We choose the camera's local 3D coordinate system as the reference frame. To convert pedestrian's GPS readings into coordinates with respect to the camera 3D coordinate frame, we need to first obtain the coordinate transformation between the world and the camera. 

\begin{figure*}
    \centering
    \includegraphics[width=1\textwidth]{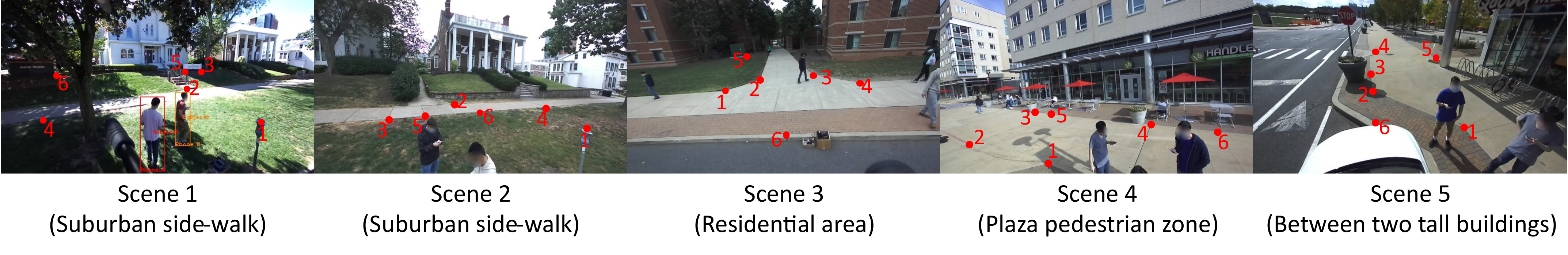}
    \caption{Experimental fields and reference points that are used to estimate world-camera transformation (Best viewed zoomed).}
    \label{fig:dataset_with_refPoints}
\end{figure*}

Consider a reference point's coordinate in GPS format (latitude, longitude, altitude), it's coordinate can be converted into 3D world Cartesian format $P = [X_W, Y_W, Z_W]$ using the WGS84 model~\cite{wgs84_wiki, Ellipsoidal_and_Cartesian, subirana2016ellipsoidal}. Its corresponding pixel coordinate on the image is $p = [u, v]$. The corresponding 3D world Cartesian coordinate and pixel coordinate have the relationship $[p,1]^\top = \mathbf{K} 
    \cdot \prescript{C}{}{\mathbf{T}}_W \cdot
    [P, 1]^\top$,
where $\mathbf{K} \in \mathbb{R}^{3\times3}$ represents the camera's intrinsics. 
$\prescript{C}{}{\mathbf{T}}_W = [\prescript{C}{}{\mathbf{R}}_W \quad \prescript{C}{}{\mathbf{t}}_W]$ is the transformation matrix from the world to camera 3D coordinate frame. It contains a rotation matrix $\prescript{C}{}{\mathbf{R}}_W \in \mathbb{R}^{3\times3}$ and a translation vector $\prescript{C}{}{\mathbf{t}}_W \in \mathbb{R}^{3\times1}$.
We adopt the AP3P~\cite{ke2017efficient} algorithm to estimate this transformation.
It takes as input 4 pairs of 3D-2D point correspondences with minimal measurement noise and outputs the estimated transformation matrix. 
As shown in Figure~\ref{fig:dataset_with_refPoints}, we collect 6 reference points' GPS coordinates at each experimental field using a survey-grade Trimble-R2 GPS receiver (meter-level accuracy). We collect the reference points' corresponding 2D pixel coordinates with a pixel information tool that displays pixel coordinates in an image that the mouse pointer is positioned over.



We implement the AP3P algorithm using OpenCV's ``solvePnP'' method. For each scene that has 6 pairs of 3D-2D reference points, we iterate through all 4-point subsets and evoke AP3P to compute the transformation matrix multiple times. We choose the transformation matrix that has the lowest re-projection error as our final estimation. 
Once the world-camera transformation $\prescript{C}{}{\mathbf{T}}_W$ is estimated, the camera-world transformation $\prescript{W}{}{\mathbf{T}}_C$ can be derived as {$\prescript{W}{}{\mathbf{T}}_C =
    \begin{bmatrix}
        \prescript{C}{}{\mathbf{R}}^\top_W & -\prescript{C}{}{\mathbf{R}}^\top_W \cdot \prescript{C}{}{\mathbf{t}}_W \\
    \end{bmatrix}$}.
From the estimated camera-world transformation matrix , we can directly obtain the estimated RSU location $P_{\text{RSU}}$ by fetching the last column of the $\prescript{W}{}{\mathbf{T}}_C$.

To evaluate the quality of the transformation matrix, we compare the estimated RSU location with the surveyed RSU location $P^*_{\text{RSU}}$ measured by Trimble R2 and compute their Euclidean distance as the RSU position error.
\begin{equation}
\small
    err_{\text{RSU}} = \| P_{\text{RSU}} - P^*_{\text{RSU}}  \|_2.
\end{equation}

We also examine the reprojection error by projecting the reference points' 3D coordinates back into the image plane using the estimated transformation matrix and compare the projected pixel coordinates with their original 2D pixel coordinates. We compute the average and the standard deviation for all reference point's reprojection error
\begin{equation}
\small
    err_{\text{reproject}} = \frac{1}{N} \sum_{i=1}^N \| \mathbf{K} \cdot \prescript{C}{}{\mathbf{T}}_W \cdot P_i - p_i \|,
\end{equation}
where $p_i$ and $P_i$ are the $i$-th reference point's pixel coordinate and 3D world coordinate respectively.


\begin{table}[]
\centering
\footnotesize
\begin{tabular}{lllll}
\hline
 & \multicolumn{1}{l}{\begin{tabular}[c]{@{}l@{}}RSU  position error (m)\end{tabular}} & \multicolumn{2}{c}{Reprojection error (pixel)} \\ 
 & & avg & std \\ \hline
Scene 1 & \multicolumn{1}{c}{1.455} & 36.3 & 46.3 \\
Scene 2 & \multicolumn{1}{c}{1.876} & 31.0 & 21.8 \\
Scene 3 & \multicolumn{1}{c}{0.886} & 27.1 & 31.2 \\
Scene 4 & \multicolumn{1}{c}{1.077} & 25.6 & 28.7 \\
Scene 5 & \multicolumn{1}{c}{1.864} & 33.5 & 38.7 \\ \hline
\end{tabular}
\caption{Roadside unit position error and reprojection error for the estimated transformation matrix in each scene. The small magnitudes of reprojection error and RSU position error suggest that the estimated transformation matrices are qualitatively satisfactory. }
\label{table: transformatino_eval}
\vspace{-4mm}
\end{table}

Table~\ref{table: transformatino_eval} shows the RSU localization error and reprojection error for 5 scenes' world-camera transformation matrices. The small magnitudes of reprojection error and RSU position error suggest that the estimated transformation matrices are qualitatively satisfactory. 

We use the above transformation matices to convert baselines' GPS coordinates into local camera coordinate frame and compare them with our GAN produced coordinates. We use localization error in average, standard deviation, median and 95 percentile as our criteria. The ground truth locations for pedestrians are their camera 3D coordinates which are derived from detected bounding boxes' centroid and depth value.

\begin{table*}[]
\footnotesize
\centering
\begin{tabular}{c|cccc|cccc|cccc}
\hline
& \multicolumn{4}{c|}{Phone GPS} & \multicolumn{4}{c|}{Phone GPS + FTM} & \multicolumn{4}{c}{\textbf{Ours (GAN generated)}} \\ 
 & avg & std & med & 95\% &    avg & std & med & 95\% &    avg & std & med & 95\%  \\ \hline
Scene 1 & 3.460 & 1.897 & 3.022 & 7.192 &   2.030 & 1.092 & 1.86 & 3.760 &   1.620 & 0.951 & 1.411 & 3.255 \\
Scene 2 & 7.314 & 4.509 & 6.155 & 17.16 &    6.055 & 2.468 & 5.637 & 10.45 &    1.822 & 1.581 & 1.311 & 5.443  \\
Scene 3 & 3.899 & 2.439 & 3.339 & 8.176 &    3.807 & 2.398 & 3.282 & 7.751 &    1.678 & 1.331 & 1.377 & 3.655  \\
Scene 4 & 3.728 & 1.552 & 3.671 & 6.538 &    2.940 & 2.289 & 2.260 & 7.443 &    1.432 & 0.927 & 1.248 & 3.175  \\
Scene 5 & 16.96 & 6.263 & 16.56 & 29.46 &    10.76 & 4.429 & 9.963 & 18.12 &    1.351 & 0.849 & 1.169 & 3.002  \\\hline
Overall & 7.443 & 6.727 & 4.773 & 21.41 &    5.339 & 4.366 & 3.923 & 13.71 &    \textbf{1.554} & \textbf{1.143} & \textbf{1.280} & \textbf{3.547}   \\ \hline
\end{tabular}
\caption{Localization error (m) in average, standard deviation, median and 95-percentile for different scenes in the test set.
}
\label{table:detailed_avg_localization_error_survey_pnp}
\vspace{-4mm}
\end{table*}



Table~\ref{table:detailed_avg_localization_error_survey_pnp} provides localization error for 5 scenes.
For baseline approaches, the phone's GPS readings have an average localization error of 7.443 m; the fused locations estimated by the particle filter that have an average localization error of 5.339 m. In comparison, the estimated coordinates generated by our proposed GAN have an average localization error of 1.554 m. 
Moreover, for baseline approaches, the phone GPS readings and particle filter estimated locations exhibit large deviations across different scenes. In scene 5, specifically, where GPS performance degrades significantly due to tall buildings, the localization errors for baseline approaches are more than 10 meter. In contrast, our method produces consistent location estimations with errors varying between 1 to 2 meters.
These comparisons suggest that our method is capable of producing location estimations that are consistently better than fused GPS location for different surrounding environments.

\subsection{Perturbation on coordinate transformation}
Readers might argue that the transformation matrix obtained by AP3P with reference points inevitably contains error due to measurement noise of the GPS collector. The true transformation could result in different localization error, which might increase the GAN estimated localization error and reduce the original GPS error. To address this concern, we conduct perturbation study on the transformation matrix and see how small perturbations affect pedestrians' localization error.

We perturb the camera-world transformation by applying a small rotation $\mathbf{R}_{p}$ and a small translation $\mathbf{t}_p$ to the original transformation.
The perturbation rotation $\mathbf{R}_p \in \mathbb{R}^{3 \times 3}$ consists of rotations with respect to the world's X, Y, and Z axis.
The rotation angle $\theta_X$, $\theta_Y$ and $\theta_Z$ are randomly drawn from a zero-mean Gaussian distribution with standard deviation $\sigma_\theta$.
The perturbation translation $\mathbf{t}_p \in \mathbb{R}^{3\times 1}$ is a zero-mean Gaussian random vector 
with covariance matrix $\mathbf{I} \cdot \sigma^2_t$.
We vary the perturbation to the transformation matrix $^W{T}_C$ by changing the value of $\sigma_\theta$ and $\sigma_t$. In evaluation we choose $\sigma_\theta = \{5\degree, 10\degree, 15\degree, 20\degree, 25\degree, 30\degree\}$ and $\sigma_t = \{0.5 \text{ m}, 1 \text{ m}, 1.5 \text{ m}, 2 \text{ m}, 2.5 \text{ m}, 3 \text{ m}\}$ respectively. For each perturbed transformation matrix we re-train our network and compute the average localization error across 5 scenes. 

\begin{table*}[]
\footnotesize
\centering
\begin{tabular}{ll|cccc|cccc|cccc}
\hline
\multicolumn{2}{c|}{Perturbation} & \multicolumn{4}{c|}{Phone GPS} & \multicolumn{4}{c|}{Phone GPS + FTM} & \multicolumn{4}{c}{\textbf{Ours (GAN generated)}} \\ 
$\sigma_\theta$ (\degree) & $\sigma_t$ (m) & avg & std & med & 95\% &    avg & std & med & 95\% &    avg & std & med & 95\%  \\ \hline
0 & 0  & 7.443 & 6.727 & 4.773 & 21.41 &    5.339 & 4.366 & 3.923 & 13.71 &    \textbf{1.554} & \textbf{1.143} & \textbf{1.280} & \textbf{3.547} \\
5 & 0.5  & 7.869 & 6.930 & 4.997 & 22.32 &    5.777 & 4.538 & 4.217 & 14.80 &    1.584 & 1.110 & 1.312 & 3.701  \\
10 & 1.0  & 7.720 & 6.508 & 5.262 & 21.25 &    5.502 & 4.268 & 4.062 & 13.63 &    1.688 & 1.242 & 1.379 & 3.994  \\
15 & 1.5  & 8.220 & 6.399 & 6.160 & 21.54 &    6.243 & 4.211 & 5.164 & 14.35 &    1.702 & 1.236 & 1.397 & 4.022  \\
20 & 2.0  & 7.692 & 5.141 & 6.272 & 16.72 &    5.956 & 3.450 & 4.914 & 12.69 &    1.930 & 1.423 & 1.560 & 4.786  \\
25 & 2.5 & 9.582 & 6.361 & 8.147 & 22.29 &    7.830 & 4.298 & 7.562 & 15.53 &    1.915 & 1.495 & 1.522 & 4.827  \\
30 & 3.0 & 9.487 & 5.530 & 7.991 & 19.62 &    7.559 & 4.055 & 6.927 & 14.88 &    2.224 & 1.842 & 1.743 & 5.799  \\ \hline
\end{tabular}
\caption{Perturbation Study results. Applying more perturbation on the transformation increases both phone GPS error and the GAN produced coordinates' localization error. But the GAN produced coordinates always exhibit much less localization error than phone GPS. }
\label{table:perturbation_study}
\vspace{-4mm}
\end{table*}

The results are shown in Table~\ref{table:perturbation_study}.
Despite the fact that perturbing the World-Camera Transformation matrix will change localization error for both original GPS and our GAN method, the relationship between them stays the same. The table shows that our proposed method always provides more than 70\% less localization error compared to smartphone's GPS readings. This suggests that the performance gain of our method is independent to the uncertainty of the transformation matrix. Considering the perfect World-Camera transformation is nearly impossible to derive, the perturbation study on the transformation matrix substantiate the argument that under most reasonable evaluation metrics, our proposed GAN estimated positions has much lower localization error than original GPS readings.

\subsection{Ablation Study}
We provide a detailed and comprehensive ablation study to explore the contribution of each data field in the phone modality. We train the GAN with different subsets and combinations of FTM, IMU, GPS and RSSI and evaluate the resulting localization error under each configuration. The results are shown in Table~\ref{table:ablation}.
In general, each individual data field makes its own contribution in training a network that produces better location estimations than the baselines.
Our proposed combination that consists of FTM, IMU and GPS provides the best localization accuracy. The other combinations, either replacing FTM with RSSI or only using a subset of the proposed combination, yield sub-optimal localization performances. 
As we remove one or more data field and reduce the dimension of input data during training phases, the localization error of the GAN produced coordinates increases during test phases. More specifically, if there is only one modality being used in the training process, the localization error goes up to 3 meters. This suggests that more information from phone modalities helps the GAN to learn a better GPS correction model that produces better localization estimations. 
The comparison in Table~\ref{table:ablation} also shows the advantage of FTM over RSSI. Unlike RSSI, which only provides the received signal strength as an indicator of the range. FTM provides a direct ranging information between the access point and a pedestrian's smartphone. This direct ranging is the key reason why the modalities that contains FTM provide lower localization error than those replaces FTM with RSSI.



\begin{table}[]
\footnotesize
\centering
\begin{tabular}{lcccc}
\hline
\begin{tabular}[l]{@{}l@{}}Phone modalities used\\ in training \end{tabular} &
\multicolumn{4}{c}{\begin{tabular}[l]{@{}l@{}}GAN produced coordinates \\localization error (m) \end{tabular}} \\ 
& avg & std & med & 95\% \\\hline
\textbf{FTM + IMU + GPS} & \textbf{1.554} & \textbf{1.143} & \textbf{1.280} & \textbf{3.547} \\ 
RSSI + IMU + GPS & 1.781 & 1.209 & 1.450 & 4.136 \\
IMU + GPS      & 2.003 & 1.325 & 1.698 & 4.558 \\
GPS          & 2.210 & 1.349 & 1.958 & 4.821 \\
FTM + IMU      & 1.708 & 1.353 & 1.376 & 4.227 \\
FTM + GPS      & 1.722 & 1.179 & 1.450 & 3.867 \\
RSSI + IMU     & 2.037 & 1.447 & 1.654 & 4.950 \\
RSSI + GPS     & 1.892 & 1.193 & 1.665 & 4.048 \\
FTM          & 3.057 & 2.216 & 2.440 & 7.578 \\
IMU          & 2.617 & 1.883 & 2.122 & 6.341 \\ \hline
\end{tabular}
\caption{Ablation Study results. The combination of FTM, IMU and GPS provides the best localization accuracy. Other combinations yield sub-optimal localization performances. }
\label{table:ablation}
\vspace{-4mm}
\end{table}


\subsection{Self-learning with associated data}
We conduct an initial evaluation on our self learning mechanism and show how it facilitate training the network at larger scale without extra data collection and manual labeling.
We first train the GAN with only one person's labeled data correspondences in the dataset. Then we deploy the network and perform inference for all the pedestrians in the training data sequences. We associate the network produced coordinates with the pedestrians' camera 3D coordinates derived from their bounding box coordinates and depth measurements to obtain more camera-phone correspondences. 
After association, we combine the associated data correspondences that belong to other pedestrians with the labeled data correspondences to form a larger training set. 
We continue to fine-tune the network using the enlarged dataset. 

Table~\ref{table:self-learning} shows the association precision and compares the localization error before and after training on the additional data correspondences that are obtained autonomously by association. 
The association precision tells us how many associated camera-phone data pairs are true correspondences. We observe that our association method provides us with a majority of good quality data correspondences.  
The gain in localization accuracy varies from 15.6\% to 26.2\%. This suggests that the additional data correspondences obtained by association are making additional contribution for the GAN to learn a better GPS correction model. These results provide a promising semi-supervised direction as the proposed association mechanism allows the network to use its output to generate more training data on its own. 
This provides an opportunity for the network to be used in more than just 5 different environment settings. With initial training using limited labeled data correspondences that are collected from one user identity at a specific scene, the proposed self-learning approach allows the network to generalize to much larger scale.

\begin{table}[]
\centering
\footnotesize
\begin{tabular}{lllll}
\hline
 & \begin{tabular}[c]{@{}l@{}}Train on one \\person's data\end{tabular} & \begin{tabular}[c]{@{}l@{}}Association\\precision\end{tabular} & \begin{tabular}[c]{@{}l@{}}Train on \\ associated\\ data\end{tabular} & \begin{tabular}[c]{@{}l@{}}\textit{Localization} \\ \textit{accuracy} \\ \textit{gain}\end{tabular} \\
 \hline
User A & 2.577 & 78.7\% & 2.176 & \textit{15.6\%} \\
User B & 2.300 & 82.2\% & 1.857 & \textit{19.3\%} \\
User C & 2.954 & 71.2\% & 2.181 & \textit{26.2\%} \\
\hline
\end{tabular}
\caption{Average localization (m) before and after training on additional data produced by association. }
\label{table:self-learning}
\vspace{-5mm}
\end{table}

\section{Conclusion}
In this paper, we propose a generative adversarial network that produces accurate location estimations based on pedestrians' phone data sequences. Trained with camera bounding boxes information and smartphone IMU, GPS and FTM measurements, our proposed GAN outperforms the phone GPS reading and a particle filter baseline with an average localization error of 1.5 m and a median localization error of 1.2 m. 
To alleviate manual labeling and data collection, we propose a self-learning approach that allows the network to use its output to generate more training data during test phases. 
By associating the produced coordinates with the coordinates from the camera-observed pedestrians, more vision-phone data correspondences can be obtained autonomously. 
Trained on the additional data correspondences, the localization accuracy of the generated coordinates is further improved by up to 26\%.


{\small
\bibliographystyle{ieee_fullname}
\bibliography{main.bib}

\begin{thebibliography}{10}\itemsep=-1pt

\bibitem{ZED_accuracy}
\url{https://support.stereolabs.com/hc/en-us/articles/206953039-How-does-the-ZED-work-}.

\bibitem{Realsense_accuracy}
\url{https://www.intel.com/content/www/us/en/support/articles/000026260/emerging-technologies/intel-realsense-technology.html}.

\bibitem{GPS_gov}
\url{https://www.gps.gov/systems/gps/performance/accuracy/}.

\bibitem{ZED}
\url{https://www.stereolabs.com/docs/object-detection/}.

\bibitem{LocationManager}
\url{https://developer.android.com/reference/android/location/LocationManager/}.

\bibitem{wgs84_wiki}
\url{https://en.wikipedia.org/wiki/World_Geodetic_System/}.

\bibitem{Ellipsoidal_and_Cartesian}
\url{https://gssc.esa.int/navipedia/index.php/Ellipsoidal_and_Cartesian_Coordinates_Conversion/}.

\bibitem{80211standard}
{"IEEE Standard for Information technology--Telecommunications and information
  exchange between systems Local and metropolitan area networks--Specific
  requirements - Part 11: Wireless LAN Medium Access Control (MAC) and Physical
  Layer (PHY) Specifications"}.
\newblock {\em {"IEEE Std 802.11-2016 (Revision of IEEE Std 802.11-2012)"}},
  pages 1--3534, Dec 2016.

\bibitem{arun2022p2slam}
Aditya Arun, Roshan Ayyalasomayajula, William Hunter, and Dinesh Bharadia.
\newblock P2slam: Bearing based wifi slam for indoor robots.
\newblock {\em IEEE Robotics and Automation Letters}, 7(2):3326--3333, 2022.

\bibitem{bao2017cvae}
Jianmin Bao, Dong Chen, Fang Wen, Houqiang Li, and Gang Hua.
\newblock Cvae-gan: fine-grained image generation through asymmetric training.
\newblock In {\em Proceedings of the IEEE international conference on computer
  vision}, pages 2745--2754, 2017.

\bibitem{campos2021orb}
Carlos Campos, Richard Elvira, Juan J~G{\'o}mez Rodr{\'\i}guez, Jos{\'e}~MM
  Montiel, and Juan~D Tard{\'o}s.
\newblock Orb-slam3: An accurate open-source library for visual,
  visual--inertial, and multimap slam.
\newblock {\em IEEE Transactions on Robotics}, 37(6):1874--1890, 2021.

\bibitem{cheng2014seamless}
Jiantong Cheng, Ling Yang, Yong Li, and Weihua Zhang.
\newblock Seamless outdoor/indoor navigation with wifi/gps aided low cost
  inertial navigation system.
\newblock {\em Physical Communication}, 13:31--43, 2014.

\bibitem{cheng2005accuracy}
Yu-Chung Cheng, Yatin Chawathe, Anthony LaMarca, and John Krumm.
\newblock Accuracy characterization for metropolitan-scale wi-fi localization.
\newblock In {\em Proceedings of the 3rd international conference on Mobile
  systems, applications, and services}, pages 233--245, 2005.

\bibitem{dellaert2012factor}
Frank Dellaert.
\newblock Factor graphs and gtsam: A hands-on introduction.
\newblock Technical report, Georgia Institute of Technology, 2012.

\bibitem{gao2021raw}
Rui Gao, Li Xu, Baocheng Zhang, and Teng Liu.
\newblock Raw gnss observations from android smartphones: Characteristics and
  short-baseline rtk positioning performance.
\newblock {\em Measurement Science and Technology}, 32(8):084012, 2021.

\bibitem{gorti2018text}
Satya~Krishna Gorti and Jeremy Ma.
\newblock Text-to-image-to-text translation using cycle consistent adversarial
  networks.
\newblock {\em arXiv preprint arXiv:1808.04538}, 2018.

\bibitem{han2018gan}
Changhee Han, Hideaki Hayashi, Leonardo Rundo, Ryosuke Araki, Wataru Shimoda,
  Shinichi Muramatsu, Yujiro Furukawa, Giancarlo Mauri, and Hideki Nakayama.
\newblock Gan-based synthetic brain mr image generation.
\newblock In {\em 2018 IEEE 15th international symposium on biomedical imaging
  (ISBI 2018)}, pages 734--738. IEEE, 2018.

\bibitem{ibrahim2018verification}
Mohamed Ibrahim, Hansi Liu, Minitha Jawahar, Viet Nguyen, Marco Gruteser,
  Richard Howard, Bo Yu, and Fan Bai.
\newblock Verification: Accuracy evaluation of wifi fine time measurements on
  an open platform.
\newblock In {\em Proceedings of the 24th Annual International Conference on
  Mobile Computing and Networking}, pages 417--427. ACM, 2018.

\bibitem{ibrahim2020wi}
Mohamed Ibrahim, Ali Rostami, Bo Yu, Hansi Liu, Minitha Jawahar, Viet Nguyen,
  Marco Gruteser, Fan Bai, and Richard Howard.
\newblock Wi-go: accurate and scalable vehicle positioning using wifi fine
  timing measurement.
\newblock In {\em Proceedings of the 18th International Conference on Mobile
  Systems, Applications, and Services}, pages 312--324, 2020.

\bibitem{ismail2019rssi}
Mohd Ismifaizul~Mohd Ismail, Rudzidatul~Akmam Dzyauddin, Shafiqa Samsul,
  Nur~Aisyah Azmi, Yoshihide Yamada, Mohd Fitri~Mohd Yakub, and Noor Azurati
  Binti~Ahmad Salleh.
\newblock An rssi-based wireless sensor node localisation using trilateration
  and multilateration methods for outdoor environment.
\newblock {\em arXiv preprint arXiv:1912.07801}, 2019.

\bibitem{ke2017efficient}
Tong Ke and Stergios~I Roumeliotis.
\newblock An efficient algebraic solution to the perspective-three-point
  problem.
\newblock In {\em Proceedings of the IEEE Conference on Computer Vision and
  Pattern Recognition}, pages 7225--7233, 2017.

\bibitem{liuvi}
Hansi Liu, Abrar Alali, Mohamed Ibrahim, Bryan~Bo Cao, Nicholas Meegan, Hongyu
  Li, Marco Gruteser, Shubham Jain, Kristin Dana, Ashwin Ashok, et~al.
\newblock Vi-fi: Associating moving subjects across vision and wireless
  sensors.

\bibitem{loeliger2004introduction}
H-A Loeliger.
\newblock An introduction to factor graphs.
\newblock {\em IEEE Signal Processing Magazine}, 21(1):28--41, 2004.

\bibitem{lu2018image}
Yongyi Lu, Shangzhe Wu, Yu-Wing Tai, and Chi-Keung Tang.
\newblock Image generation from sketch constraint using contextual gan.
\newblock In {\em Proceedings of the European conference on computer vision
  (ECCV)}, pages 205--220, 2018.

\bibitem{mao2017least}
Xudong Mao, Qing Li, Haoran Xie, Raymond~YK Lau, Zhen Wang, and Stephen
  Paul~Smolley.
\newblock Least squares generative adversarial networks.
\newblock In {\em Proceedings of the IEEE international conference on computer
  vision}, pages 2794--2802, 2017.

\bibitem{misra1999gps}
Pratap Misra, Brian~P Burke, and Michael~M Pratt.
\newblock Gps performance in navigation.
\newblock {\em Proceedings of the IEEE}, 87(1):65--85, 1999.

\bibitem{mur2015orb}
Raul Mur-Artal, Jose Maria~Martinez Montiel, and Juan~D Tardos.
\newblock Orb-slam: a versatile and accurate monocular slam system.
\newblock {\em IEEE transactions on robotics}, 31(5):1147--1163, 2015.

\bibitem{NEURIPS2019_9015}
Adam Paszke, Sam Gross, Francisco Massa, Adam Lerer, James Bradbury, Gregory
  Chanan, Trevor Killeen, Zeming Lin, Natalia Gimelshein, Luca Antiga, Alban
  Desmaison, Andreas Kopf, Edward Yang, Zachary DeVito, Martin Raison, Alykhan
  Tejani, Sasank Chilamkurthy, Benoit Steiner, Lu Fang, Junjie Bai, and Soumith
  Chintala.
\newblock Pytorch: An imperative style, high-performance deep learning library.
\newblock In H. Wallach, H. Larochelle, A. Beygelzimer, F. d\textquotesingle
  Alch\'{e}-Buc, E. Fox, and R. Garnett, editors, {\em Advances in Neural
  Information Processing Systems 32}, pages 8024--8035. Curran Associates,
  Inc., 2019.

\bibitem{qin2018vins}
Tong Qin, Peiliang Li, and Shaojie Shen.
\newblock Vins-mono: A robust and versatile monocular visual-inertial state
  estimator.
\newblock {\em IEEE Transactions on Robotics}, 34(4):1004--1020, 2018.

\bibitem{rostami2019light}
Ali Rostami, Bin Cheng, Hongsheng Lu, John~B Kenney, and Marco Gruteser.
\newblock A light-weight smartphone gps error model for simulation.
\newblock In {\em 2019 IEEE 90th Vehicular Technology Conference
  (VTC2019-Fall)}, pages 1--5. IEEE, 2019.

\bibitem{shi2019generative}
Yi Shi, Kemal Davaslioglu, and Yalin~E Sagduyu.
\newblock Generative adversarial network for wireless signal spoofing.
\newblock In {\em Proceedings of the ACM Workshop on Wireless Security and
  Machine Learning}, pages 55--60, 2019.

\bibitem{shi2020generative}
Yi Shi, Kemal Davaslioglu, and Yalin~E Sagduyu.
\newblock Generative adversarial network in the air: Deep adversarial learning
  for wireless signal spoofing.
\newblock {\em IEEE Transactions on Cognitive Communications and Networking},
  7(1):294--303, 2020.

\bibitem{subirana2016ellipsoidal}
J~Sanz Subirana, JM~Juan Zornoza, and M Hern{\'a}ndez-Pajares.
\newblock Ellipsoidal and cartesian coordinates conversion, 2016.

\bibitem{wang2017woloc}
Jin Wang, Nicholas Tan, Jun Luo, and Sinno~Jialin Pan.
\newblock Woloc: Wifi-only outdoor localization using crowdsensed hotspot
  labels.
\newblock In {\em IEEE INFOCOM 2017-IEEE Conference on Computer
  Communications}, pages 1--9. IEEE, 2017.

\bibitem{zhang2018quality}
Xiaohong Zhang, Xianlu Tao, Feng Zhu, Xiang Shi, and Fuhong Wang.
\newblock Quality assessment of gnss observations from an android n smartphone
  and positioning performance analysis using time-differenced filtering
  approach.
\newblock {\em Gps Solutions}, 22(3):1--11, 2018.

\end{thebibliography}
}

\end{document}